%% file: iclr2026_conference.tex
\documentclass{article} 
\usepackage[dvipsnames,table]{xcolor}         
\usepackage{iclr2026_conference,times}

\input{math_commands.tex}

\usepackage{booktabs}       
\usepackage{graphicx}
\usepackage{hyperref}
\usepackage{url}
\usepackage{xspace}
\usepackage{multirow}
\usepackage{listings}
\usepackage{stfloats}

\usepackage{booktabs}
\usepackage{colortbl}
\usepackage{wrapfig} 
\usepackage{enumitem}

\definecolor{b1}{HTML}{1280B0}
\definecolor{b2}{HTML}{25537D}
\colorlet{punct}{red!60!black}
\definecolor{background}{HTML}{EEEEEE}
\definecolor{delim}{RGB}{20,105,176}
\colorlet{numb}{magenta!60!black}

\newcommand{\name}{\textsc{TreeRPO}\xspace}

\title{\name: Tree Relative Policy Optimization}

\author{
~~~~~~~~~~~~~~~~~~~~Zhicheng Yang\textsuperscript{1}~~
Zhijiang Guo\textsuperscript{1,2}~~
Yinya Huang\textsuperscript{3}~~
Yongxin Wang\textsuperscript{6}~~
\\
~~~~~~~~~~~~~~~~~~~~~~~~~~~~~~~~~~~~~~~~~\textbf{Yiwei Wang\textsuperscript{4}~~Xiaodan Liang\textsuperscript{5,6}~~Jing Tang\textsuperscript{1,2}}\thanks{Corresponding author: Jing Tang.} \\
~~~~~~~~~~~~~~~~~~~~~$^1$The Hong Kong University of Science and Technology (Guangzhou) \\
~~~~~~$^2$The Hong Kong University of Science and Technology ~~
$^3$ETH AI Center, ETH Zurich ~~ \\
~~~~~~~~~~~~~~~~~~$^4$University of California, Merced ~~
$^5$Sun Yat-sen University ~~
$^6$MBZUAI ~~\\
~~~~~~~~~~~~~~~~~~~~~~~~~~~~~~~~~~~~~~~~~~~~~~~~~~~~~~~~\texttt{yangzhch6@gmail.com} \\ \\
~~~~~~~~~~~~\textit{Project Repo}: \href{https://github.com/yangzhch6/Mirror-Critique}{ \ttfamily https://github.com/yangzhch6/Mirror-Critique}
}

\usepackage[most,skins,theorems]{tcolorbox}
\tcbset{
  aibox/.style={
    width=\linewidth,
    top=7pt,
    bottom=2pt,
    colback=blue!6!white,
    colframe=black,
    colbacktitle=black,
    enhanced,
    center,
    attach boxed title to top left={yshift=-0.1in,xshift=0.15in},
    boxed title style={boxrule=0pt,colframe=white,},
  }
}
\newtcolorbox{AIbox}[2][]{aibox,title=#2,#1}

\newcommand{\highlight}[1]{{\color{red!20!violet}#1}}

\definecolor{table-blue}{RGB}{173, 216, 230}
\definecolor{darkgreen}{rgb}{0.0, 0.5, 0.0}
\definecolor{darkblue}{rgb}{0, 0, 0.5}
\definecolor{paleviolet}{HTML}{E1EEFC}

\iclrfinalcopy 
\begin{document}

\maketitle

\begin{abstract}
Large Language Models (LLMs) have shown remarkable reasoning capabilities through Reinforcement Learning with Verifiable Rewards (RLVR) methods. However, a key limitation of existing approaches is that rewards defined at the full trajectory level provide insufficient guidance for optimizing the intermediate steps of a reasoning process. To address this, we introduce \textbf{\name}, a novel method that estimates the mathematical expectations of rewards at various reasoning steps using tree sampling. Unlike prior methods that rely on a separate step reward model, \name directly estimates these rewards through this sampling process. Building on the group-relative reward training mechanism of GRPO, \name innovatively computes rewards based on step-level groups generated during tree sampling. This advancement allows \name to produce fine-grained and dense reward signals, significantly enhancing the learning process and overall performance of LLMs. Experimental results demonstrate that our \name algorithm substantially improves the average Pass@1 accuracy of Qwen-2.5-Math on test benchmarks, increasing it from 19.0\% to 35.5\%. Furthermore, \name significantly outperforms GRPO by 2.9\% in performance while simultaneously reducing the average response length by 18.1\%, showcasing its effectiveness and efficiency.


\begin{figure}[htbp] 
    \centering
    \includegraphics[width=0.8\linewidth]{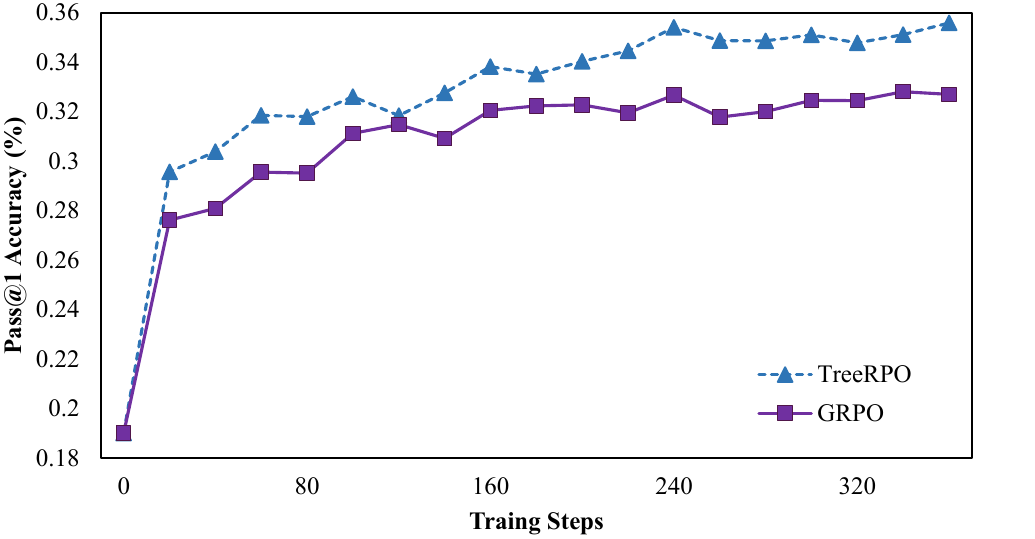}
    \caption{The average Pass@1 accuracy of \name and GRPO with Qwen-2.5-Math-1.5b on four mathematical benchmarks: MATH-500, OlympiadBench, Minerva, and AIME.}\label{fig:abstract}
\end{figure}

\end{abstract}

\section{Introduction}
Recent advancements in test-time scaling with reinforcement learning methods bring milestone progress to Large Language Models (LLMs). Reasoning models such as OpenAI O1~\citep{o1}, DeepSeek R1~\citep{guo2025deepseek}, and QwQ~\citep{qwq} have demonstrated significantly superior performance in complex reasoning tasks. Reinforcement Learning with Verifiable Rewards (RLVR) plays a pivotal role in this progress, which enhances the model’s performance by continuously exploring reasoning paths toward correct answers on verifiable problems.

In the realm of LLM-RL integration for complex reasoning, existing approaches can be broadly categorized into two paradigms: \textit{reward model-based} methods~\citep{rlhf,rlhf-sw,ppo} and \textit{reward model-free} methods~\citep{dpo,deepseekmathgrpo,simple-rl,deepscaler2025}. Among reward model-based methods, reward models are typically divided into outcome reward models (ORMs;~\citealt{cobbe2021training,ovm}) and process reward models (PRMs;~\citealt{verify_stepbystep,mathshepherd, autopsv,chen2025spcevolvingselfplaycritic}). ORMs provide a single scalar reward for the entire generation sequence, while PRMs offer step-wise evaluations of the reasoning path. The fine-grained, dense reward signals from PRMs generally yield superior RL performance compared to ORMs. However, acquiring high-quality training data for PRMs remains challenging, as accurately annotating the correctness of intermediate reasoning steps requires substantial domain expertise. This data scarcity significantly hinders the scalability of PRM-based approaches.

Recent breakthroughs in enhancing LLM reasoning capabilities, such as GRPO~\citep{deepseekmathgrpo} and its variants~\citep{dapo,vapo}, have adopted a reward model-free paradigm. These methods leverage verifiable reward functions trained on complex reasoning datasets, where rewards are determined by whether the model's final output matches the ground-truth numerical answer or passes predefined unit tests in programming tasks. This approach achieves remarkable scalability by eliminating the need for human annotations or reward models. However, similar to ORMs, these rule-based methods only provide trajectory-level rewards, offering limited guidance for optimizing intermediate reasoning steps. Consequently, the question of how to deliver dense, fine-grained reward signals without relying on reward models presents an important research direction.

To address this challenge, we propose \textbf{\name}, a novel approach that estimates the mathematical expectations of rewards at various reasoning steps through tree sampling. Unlike previous methods that require explicit step-level reward models, \name employs a tree-based sampling mechanism to approximate these expectations. Building upon GRPO's group-relative reward training framework, \name innovatively computes rewards based on step-level groups within the sampled tree structure. This design enables the generation of fine-grained, dense reward signals that guide the model's reasoning process more effectively while maintaining the scalability advantages of verifiable reward functions. Through this approach, \name achieves more efficient and effective optimization of LLM reasoning capabilities.

To summarize, our main contributions are as follows:
\begin{itemize}

\item To the best of our knowledge, \name is the first reward model-free method that provides dense process reward signals through tree sampling and group relative reward computation, significantly enhancing the efficiency of RL-based reasoning optimization.


\item Through extensive experimentation, \name was found to significantly increase Qwen-2.5-Math-1.5B's Pass@1 accuracy on various benchmarks from 19.0\% to 35.5\%, including a \textbf{2.9\%} lead over GRPO.

\item Detailed analysis demonstrates that \name achieves higher accuracy and reduces token consumption. Specifically, it cuts the average response length on test benchmarks by \textbf{18.1\%} compared to GRPO, showcasing more efficient and precise reasoning.

\end{itemize}

\section{Related Works}

\subsection{Eliciting Complex Reasoning Ability}
Complex reasoning tasks~\citep{hendrycksmath2021,olympiadbench,minerva,MrBen, yang2025optibench,xiang2025seephysdoesseeinghelp} such as mathematical problem solving are one of the most challenging tasks for LLMs. Various methods are proposed to elicit the reasoning ability of LLMs. These approaches can be divided into two groups: \\
1) \textit{In-context learning}: These methods aim to improve the reasoning ability of LLMs by designing various prompting strategies and frameworks without updating the model parameters. Chain-of-thought (CoT;~\citealt{wei2022cot}) prompting shows that intermediate reasoning steps can greatly improve model performance. Subsequent research~\citep{zhang2023cumulative,treeofthoughts,bi2023program,alignedcot} has further enhanced CoT through various methods.\\
2) \textit{Fine-tuning}: This line of approaches~\citep{logicsolver,yu2024metamath,mathgenie,efficode,tong2024dartmath} involve finetuning on extensive and high-quality datasets to improve reasoning capabilities. The core of these methods is to construct high-quality question-response pairs with chain-of-thought reasoning processes. MetaMath~\citep{yu2024metamath} focuses on data augmentation for both questions and answer texts. MathGenie~\citep{mathgenie} collects a vast amount of data through open-source language models. DART-Math~\citep{tong2024dartmath} generates diverse solutions with the difficulty-aware rejection sampling.
Recent studies~\citep{deepseekmathgrpo,hu2025openreasonerzeroopensourceapproach,simple-rl,deepscaler2025,dapo,vapo} have explored reinforcement learning in complex reasoning tasks and have acquired great achievements.
Inspired by recent successes in reinforcement learning for complex reasoning tasks, we propose \name, an innovative reinforcement learning method that leverages tree sampling to further enhance LLM reasoning ability.

\subsection{Reinforcement Learning with LLMs}

Reinforcement Learning from Human Feedback (RLHF;~\citealt{rlhf}) has been widely used in LLM alignments. Direct Preference Optimization (DPO;~\citealt{dpo}) is further proposed to simplify the training pipeline of RLHF, which directly uses pair-wise preference data for model optimization. Recent studies~\citep{o1,guo2025deepseek,grok3,gemini-thinking,qwq,k1.5} have shown that reinforcement learning can significantly improve the reasoning ability of models. This type of work can roughly be divided into two categories: \\
1) \textit{Reward model-based}: There are two primary types of reward models: the Outcome Reward Model (ORM) and the Process Reward Model (PRM). Prior effort~\citep{verify_stepbystep} suggests that PRM outperforms ORM due to the fine-grained step-by-step reward signals. Math-Shepherd~\citep{mathshepherd} trains a PRM by estimating the potential for a given reasoning step. However, training a reward model requires extensive, high-quality annotated data, especially for PRMs. This hinders the scaling of reward models in the field of complex reasoning.\\
2) \textit{Reward model-free}: DPO is one of these methods, but it requires the elaborate construction of pairwise data for training.
Step-DPO~\citep{stepdpo} constructs a pipeline to generate pair-wise step-level data and surpasses the performance of DPO.
The other line of research~\citep{deepseekmathgrpo,hu2025openreasonerzeroopensourceapproach,simple-rl,deepscaler2025} has shown that verification functions are effective in improving the reasoning capabilities of LLMs. 
They avoid the need for reward models, offering a simple yet effective approach. The typical methods are GRPO~\citep{deepseekmathgrpo} and its variants DAPO~\citep{dapo} and VAPO~\citep{vapo}. However, rule-based reward is similar to ORM, providing trajectory-level reward signals rather than fine-grained process reward signals. 
VinePPO \citep{kazemnejad2025vineppo} conduct value estimation with Monte Carlo Tree Search. However, the auxiliary rollouts are not used in policy gradient updates, while our TreeRPO utilize all rollouts of the whole tree. 
Concurrently, SPO \citep{guo2025segmentpolicyoptimizationeffective} also leverages segment-level advantage estimation at an intermediate granularity, achieving a better balance by offering more precise credit assignment.
In general, unlike existing efforts, \name achieves fine-grained, dense reward signals without relying on a separate reward model. \name can offer a more scalable solution for obtaining dense reward signals in complex reasoning tasks.

\section{\name: Methodology}
In this section, we elaborate on the proposed \name.
First, we present tree sampling in Section~\ref{sec:tree_sampling}, which is designed to construct step step-level group to enhance long-chain reasoning abilities with GRPO.
Next, in Section~\ref{sec:pruning}, we introduced the pruning strategy to improve the sampling and training efficiency in \name.
In Section~\ref{sec:adv}, we discuss the numerical influence of standardized binary rewards and continuous rewards on advantage computation and propose a new advantage computation method for continuous rewards.

\begin{figure}[htbp] 
    \centering
    \includegraphics[width=0.99\linewidth]{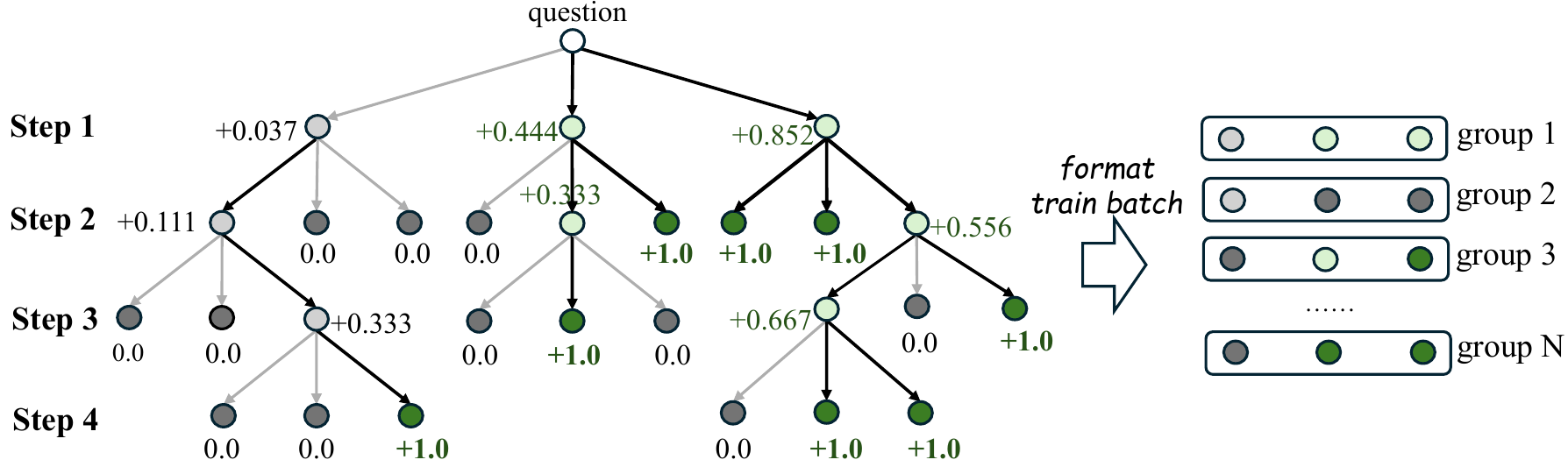}
    \caption{The sampling process of our \name. \name starts from the question, sampling $N$ nodes at each step until generation is completed or the maximum depth limit $D$ is reached. Then, a verifiable reward function is used to evaluate all leaf nodes and then back-propagates the rewards to their parent nodes, thereby obtaining intermediate step rewards, which achieves process reward signaling. We traverse each node and aggregate all children steps of a node into a group to compute advantages, which are finally formatted into the training batch.}\label{fig:framework}
\end{figure}

\subsection{Tree Sampling}
\label{sec:tree_sampling}
While GRPO has been proven to be effective and suitable for scaling in complex reasoning tasks with verifiable reward, it only provides the trajectory-level reward by evaluating the final answer of the generated sequences. Instead, to provide step-level reward estimation without using a reward model, we designed tree sampling.

Given an input question \( q \), the language model generates an \( N \)-ary tree through iterative sampling, governed by the following constraints:

\begin{itemize}
    \item \textbf{Branching Factor}: At each decoding step, the model samples \( N \) candidate continuations, expanding \( N \) new branches from the current node.
    \item \textbf{Depth Limit}: The tree expansion terminates when any path reaches the maximum depth \( D \), ensuring tractability.
    \item \textbf{Step Segmentation}: We directly divide the steps according to the token length. Each step produces at most \( L_{\text{step}} \) tokens per branch. Generation halts for a branch if a stop token is generated, or the branch violates reaches depth limit. A more precise step division method is our future work.
\end{itemize}

As shown in Figure~\ref{fig:framework}, the tree's reward computation follows a bottom-up recursive expectation scheme, where:

\begin{itemize}
    \item \textbf{Leaf Evaluation}:  
    For each leaf node \( v_{\text{leaf}} \), the verification function \( \phi \) takes the \textit{entire path} \( P = [v_{\text{root}}, \dots, v_{\text{leaf}}] \) as input and computes the reward:
    \[
    r_{\text{leaf}} = \phi(P) = \phi\big([v_{\text{root}}, \dots, v_{\text{leaf}}]\big),
    \]

    \item \textbf{Parent Propagation}:  
    Non-leaf nodes aggregate rewards from their children:
    \[
    r_{\text{node}} = \mathbb{E}_{c \in \text{Children}(v_{\text{node}})} \big[ r_c \big].
    \]
    This propagates bottom-up, weighting all viable completion paths.
    
\end{itemize}

In conclusion, our tree sampling framework estimates the reward of each step as its potential to deduce the correct final answer.

\subsection{Data Pruning}
\label{sec:pruning}
Similar to the Dynamic Sampling strategy of DAPO, we filter out the samples to keep all data samples in the training batch with effective gradients.

In the data construction pipeline of \textsc{TreeRPO}, a \textbf{group} $\mathcal{G}$ is formally defined as the set of child nodes ${c_1, \ldots, c_n}$ originating from a common parent node $p$, as illustrated in Figure~\ref{fig:framework}. Adopting a strategy analogous to the dynamic sampling approach in \textsc{DAPO}, we perform group-level filtering based on reward distribution characteristics.

\begin{equation}
\Delta R_{\mathcal{G}} = \max_{c_i \in \mathcal{G}} R(c_i) - \min_{c_j \in \mathcal{G}} R(c_j)
\end{equation}

where $R(c_i)$ denotes the reward associated with child node $c_i$. We introduce a variance threshold $\tau$ such that a group $\mathcal{G}$ is included in the training batch $\mathcal{B}$ if and only if:

\begin{equation}
\mathcal{G} \in \mathcal{B} \iff \Delta R_{\mathcal{G}} > \tau
\end{equation}

The threshold $\tau$ operates as a hyperparameter controlling the trade-off between sample diversity and learning signal strength in the batch construction process.

This data selection criterion ensures all samples in the batch with effective gradients and improves the efficiency of the training process.

\subsection{Advantage Computation}
\label{sec:adv}
In the vanilla GRPO framework, the advantage estimation is derived by normalizing binary rewards:
\begin{equation}
\hat{A}_{i,t} = \frac{r_i - \text{mean}(\{R_i\}_{i=1}^G)}{\text{std}(\{R_i\}_{i=1}^G)}.
\label{eq:grpo_adv}
\end{equation}

However, when applied to continuous rewards, this approach introduces significant bias. For instance, two reward sequences, $\mathbf{R}_1 = [0, 0, 1, 1]$ and $\mathbf{R}_2 = [0.49, 0.49, 0.51, 0.51]$, produce identical normalized advantages $[-1, -1, 1, 1]$, despite their distinct reward distributions. While $\mathbf{R}_1$ exhibits a clear bimodal separation, $\mathbf{R}_2$ contains only minor variations (a maximal difference of $0.02$). This indicates that standard normalization fails to properly scale advantages for continuous rewards, leading to misleading policy updates.

To mitigate this bias, we propose an alternative advantage computation that preserves the statistical properties of binary reward normalization while accommodating continuous rewards. Instead of computing the empirical variance from $\mathbf{R}$, we define the normalization factor as $\sigma = \mu(1 - \mu)$, where $\mu$ is the mean reward. This formulation maintains consistency with the variance of Bernoulli-distributed rewards ($\text{Var}[R] = \mu(1 - \mu)$) while generalizing to continuous settings.

For a given reward sequence $\mathbf{R} = [R_1, R_2, \dots, R_n]$, the advantage is computed as:

\begin{equation}
\begin{aligned}
\mu = \frac{1}{n}\sum_{i=1}^n &R_i ~~~~~~\sigma = \mu(1 - \mu) \\
A_i &= \frac{R_i - \mu}{\sigma}
\end{aligned}
\end{equation}

By fixing the variance term $\sigma$ to $\mu(1 - \mu)$, we ensure that advantage values remain interpretable and stable, avoiding the overamplification of small differences in continuous rewards. This approach bridges the gap between binary and continuous reward normalization while maintaining the original scaling behavior of GRPO.

\subsection{Objective of \name}
We adopt the clipped objective of GRPO, together with a directly imposed KL penalty term:
Additionally, the KL-regularization between current policy $\pi_\theta$ and the reference policy $\pi_\text{ref}$ is directly added to the loss function:
\begin{footnotesize}
\begin{equation}
\begin{aligned}
\mathcal{J}_\text{TreeRPO}&(\theta) = \mathbb{E}_{(q\sim \mathcal{D}, \{o_i\}_{i=1}^\mathcal{G}\sim \pi_{\theta_\text{old}}(q)} \\&
\Bigg[ \frac{1}{G}\sum_{i=1}^{G} \frac{1}{|o_i|}\sum_{t=1}^{|o_i|} \Bigg( 
\min \Big( r_{i,t}(\theta) \hat{A}_{i,t},  
\ \text{clip} \Big( r_{i,t}(\theta), 1 - \varepsilon, 1 + \varepsilon \Big) \hat{A}_{i,t} \Big)
- \beta D_{\text{KL}}(\pi_{\theta} || \pi_{\text{ref}}) 
\Bigg) \Bigg],
\label{eq:grpoloss}
\end{aligned}
\end{equation}
\end{footnotesize}
where
\begin{equation}
    r_{i,t}(\theta)=\frac{\pi_{\theta}(o_{i,t} \mid q, o_{i,<t})}{\pi_{\theta_{\text{old}}}(o_{i,t} \mid q,o_{i,<t})}.
\end{equation}

\section{Experiments}

\paragraph{Datasets.} We conduct the evaluation of our experiments using 4 widely used mathematical reasoning benchmarks: MATH-500~\citep{verify_stepbystep}, OlympiadBench~\citep{olympiadbench}, MinvervaMath~\citep{minerva}, and AIME24. Among them, Math-500 are 500 items screened out from the original MATH test split. The subset consists of 500 representative problems, and the evaluation produces similar results to the full-set evaluation. In the training scenario, we use the training split of MATH dataset, which contains 7.5k high-quality training samples. In the future, we will extend the experiment to the DeepScaler~\citep {deepscaler2025} training data, which is a more challenging dataset for mathematical reasoning.

\paragraph{Parameter Setting.}
Our experiments are based on Qwen2.5-Math series language models~\citep{qwen25math}.
In the evaluation procedure, we set the temperature as 0.6 to sample 8 candidate responses for each question.
In the reinforcement learning training procedure, we set the temperature as 0.6 and roll out 8 responses for each question. The learning rate is 1e-6 for both GRPO and \name.
The coefficients for KL divergence and entropy loss are $\beta=0.001$ and $\alpha=-0.001$, respectively. For GRPO, the training batch size is 128 and the mini-batch size is 64. For our \name, the training batch size is 128. Since the training data size of each step of \name is floating, the size of our mini-batch is obtained as half of the training data size. 
By default, the maximum prompt length is 512, and the maximum response length is 1152 for GRPO.
For \name, the maximum prompt length is 512, the maximum step length $L_{step}$ is 384, the maximum depth $D$ of tree sampling is set as 3, and the $N$-ary is set as 8. For better efficiency, we set the data pruning coefficient $\tau$ to 0.1 as described in Sec. \ref{sec:pruning}.

\paragraph{Implementation Details.}
We follow the rllm~\citep{deepcoder2025,deepscaler2025} framework which is derived from the verl~\citep{verl} pipeline. Both rllm and verl integrate the vllm~\citep{vllm} framework for efficient inference of models.
All of our experiments are conducted on A800 GPUs. At present, the LLM of our experiment is the Qwen2.5-Math series. Due to the limitations of time and computation resources, we have only reported the 1.5b model. We plan to conduct experiments on 7b and 32b as soon as possible

\paragraph{Metrics.}
We use the same verification function in rllm to evaluate the performance of LLMs. Compared with other repositories, the reward function implemented by rllm is more complete and systematic. For the test results, the accuracy rate we report is \textbf{pass@1(avg@8)} performance for all tested benchmarks.

\begin{table}[htbp]
    \centering
    \renewcommand\arraystretch{1.3}
    \setlength{\tabcolsep}{5pt}
    \caption{Overall performance of \textit{Pass@1} (\textit{Avg@16}) performance of Qwen2.5-Math series.}
    \begin{tabular}{l|c|c|c|c|c}
       \toprule
       \textbf{Method} & AIME24 & MATH500 & Olympiad & Minerva & Macro Accuracy \\
       \midrule
       \rowcolor{gray!16} \multicolumn{6}{c}{\textit{Qwen2.5-Math-\highlight{1.5B} as the Base Model}} \\
        GRPO Baseline & 13.8 & 67.9 & 28.5 & 20.5 & 32.7 \\
        TreeRPO & 16.8 ({\color{darkgreen!100}$\uparrow$\textbf{+3.0}}) & 70.7 ({\color{darkgreen!100}$\uparrow$\textbf{+2.8}}) & 30.9 ({\color{darkgreen!100}$\uparrow$\textbf{+2.6}}) & 24.0 ({\color{darkgreen!100}$\uparrow$\textbf{+3.5}}) & 35.6 ({\color{darkgreen!100}$\uparrow$\textbf{+2.9}}) \\
        \midrule
        \rowcolor{gray!16} \multicolumn{6}{c}{\textit{Qwen2.5-Math-\highlight{7B} as the Base Model}} \\
        GRPO Baseline & 26.7 & 74.3 & 34.7 & 27.1 & 40.7 \\
        TreeRPO & 26.7 & 75.5 ({\color{darkgreen!100}$\uparrow$\textbf{+1.2}}) & 35.4 ({\color{darkgreen!100}$\uparrow$\textbf{+0.7}}) & 28.1 ({\color{darkgreen!100}$\uparrow$\textbf{+1.0}}) & 41.4 ({\color{darkgreen!100}$\uparrow$\textbf{+0.7}}) \\
        \toprule
    \end{tabular}
    \label{tab:main}
\end{table}

\subsection{Main Results}

We show the performance comparison of GRPO baseline and our TreeRPO on Qwen2.5-Math-1.5/7B in four selected benchamrk: AIME24, MATH-500, Olympiad Benchamrk, and Minerva Math. As illustrated in Table \ref{tab:main}, for Qwen2.5-Math-1.5B, the Macro Accuracy has improved by 2.9\%. Furthermore, we consider that the reason why the improvement and repetition of Qwen2.5-Math-7B is not as significant as that of Qwen2.5-Math-1.5B lies in the fact that the MATH training data for Qwen2.5-Math-7B is too simple, resulting in the improvement of the algorithm not being significantly reflected. In generall, our \textbf{TreeRPO} has achieved a consistency improvement compared to GRPO baseline.

\begin{figure}[htbp] 
    \centering
    \includegraphics[width=0.99\linewidth]{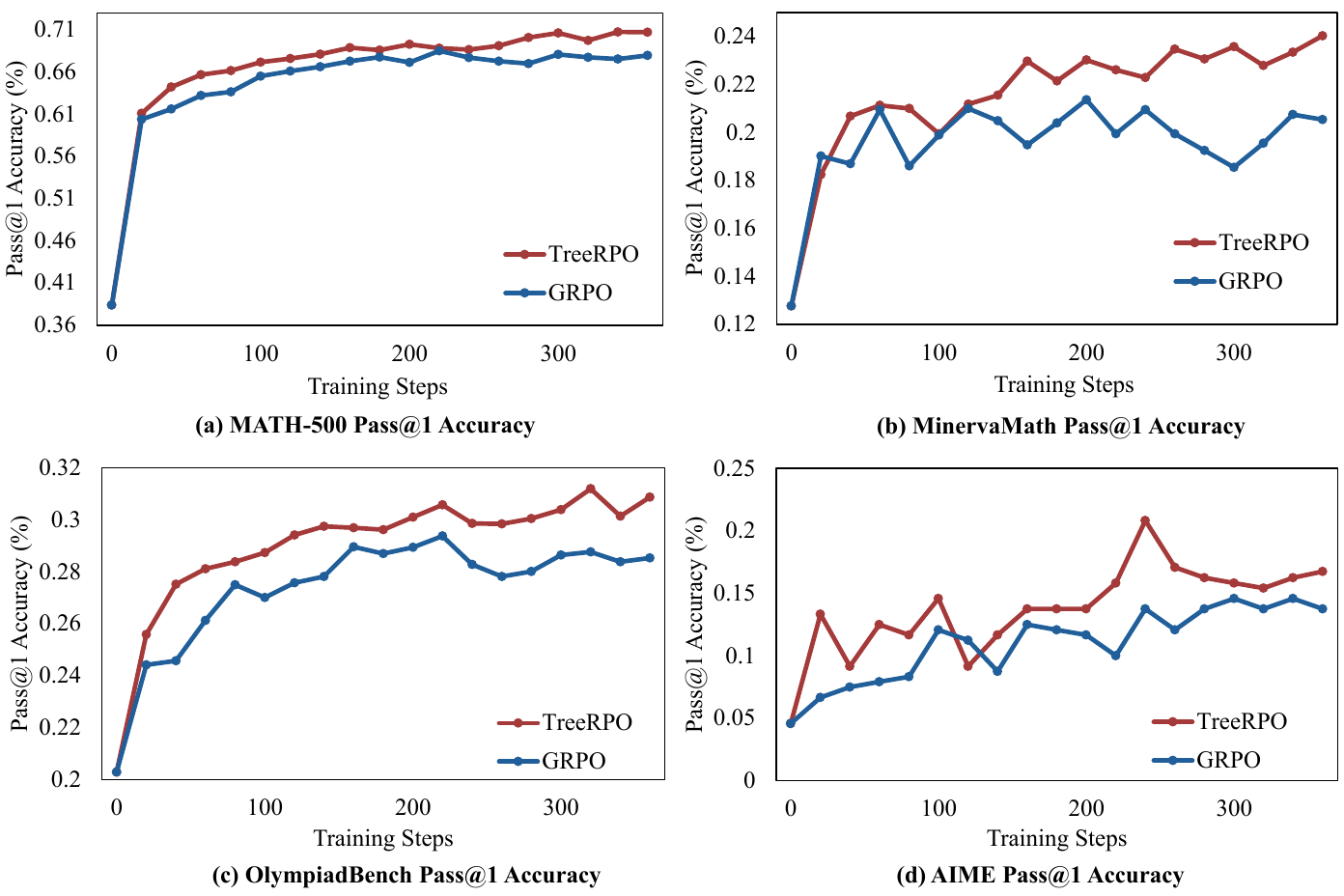}
    \caption{Performance comparison of our \name and GRPO on the four selected benchmarks: Math-500, MinervaMath, OlympiadBench, and AIME. The experiments are conducted with Qwen2.5-Math-1.5b, an LLM pretrained with a large amount of mathematical corpus.}\label{fig:performance-bs128}
\end{figure}

\begin{figure}[htbp] 
    \centering
    \includegraphics[width=0.99\linewidth]{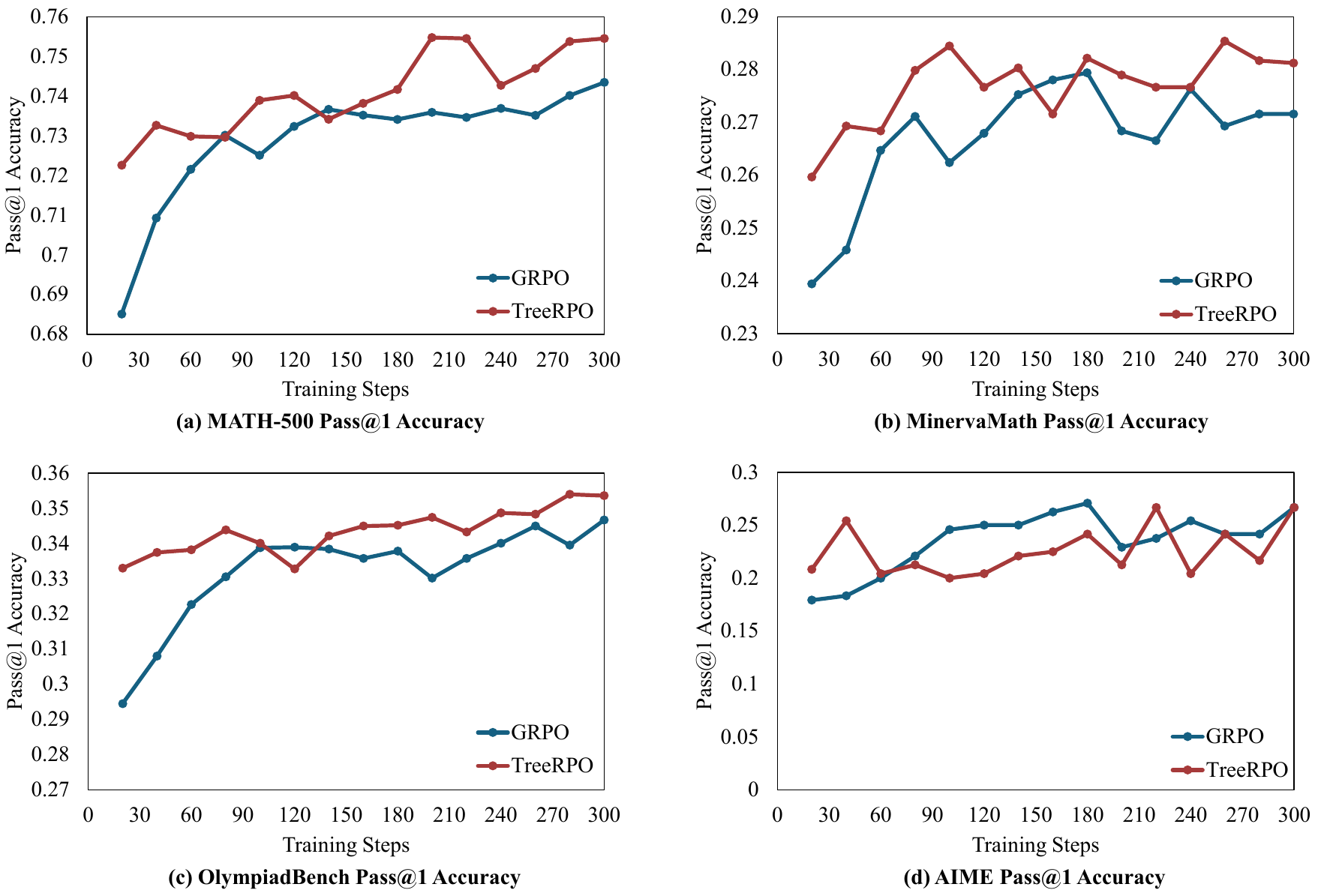}
    \caption{Performance comparison of our \name and GRPO on the four selected benchmarks: Math-500, MinervaMath, OlympiadBench, and AIME. The experiments are conducted with Qwen2.5-Math-1.5b.}\label{fig:performance-7b}
\end{figure}

\paragraph{\name demonstrates significant performance improvements.}
We conduct \name and GRPO on Qwen2.5-Math-1.5b model with the training split of the MATH dataset, and conduct the evaluation on four selected benchmarks: Math-500, MinervaMath, OlympiadBench, and AIME.
As shown in Figure~\ref{fig:performance-bs128}, our \name outperform GRPO on all of the tested benchmarks. We further show the dynmaic results of Qwen2.5-Math-7B in Figure \ref{fig:performance-7b}.
After training 300 steps for Qwen2.5-Math-1.5B, our \name outperforms GRPO by 2.7\% on MATH-500, 3.5\% on MinervaMath, 2.4\% on OlympiadBench, and 3.0\% on AIME, respectively. As illustrated in Figure~\ref{fig:abstract}, \name outperforms the overall performance of GRPO by \textbf{2.9\%}.
In conclusion, \name has demonstrated consistent superiority on multiple benchmarks.

\begin{figure}[htbp] 
    \centering
    \includegraphics[width=0.99\linewidth]{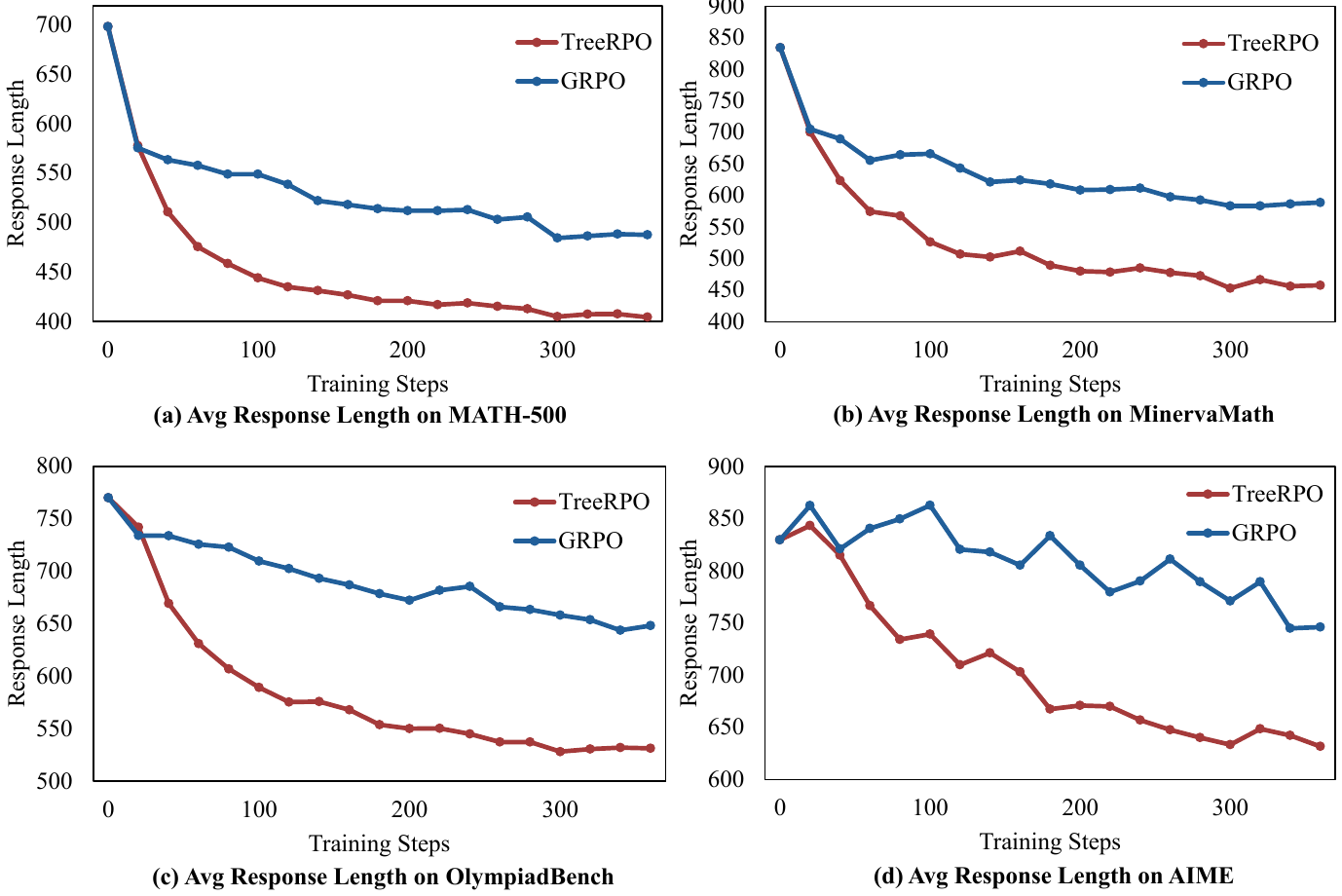}
    \caption{Response Length comparison of our \name and GRPO on the four selected benchmarks: Math-500, MinervaMath, OlympiadBench, and AIME. The experiments is conducted with Qwen2.5-Math-1.5b}\label{fig:length-bs128}
\end{figure}

\begin{figure}[t] 
    \centering
    \includegraphics[width=0.99\linewidth]{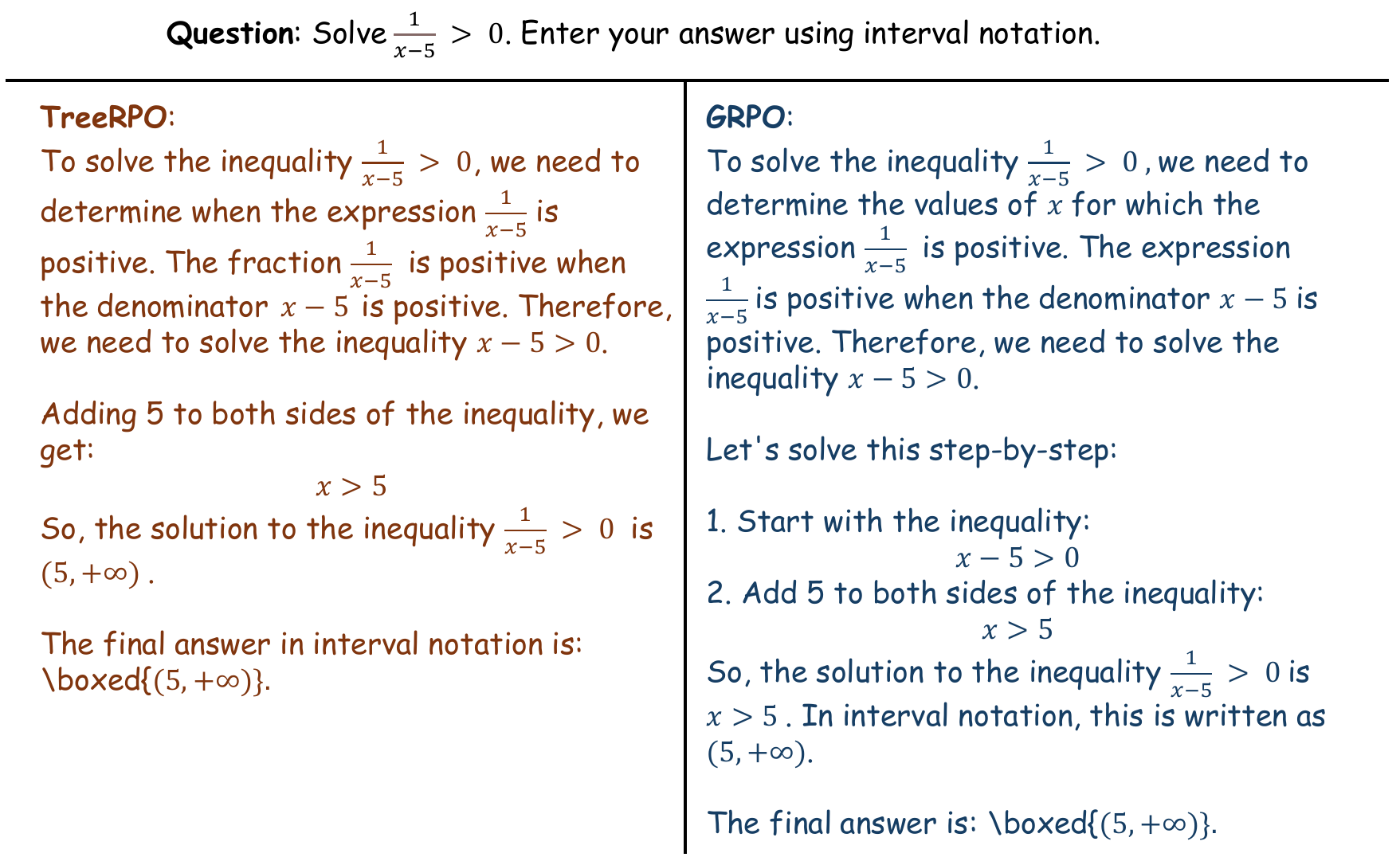}
    \caption{Responses of \name and GRPO of a simple question.}\label{fig:simple_case}
    \vspace{-4mm}
\end{figure}

\paragraph{\name demonstrates efficiency advantage in token usage.}
We conduct \name and GRPO on the Qwen2.5-Math-1.5b model with the training split of the MATH dataset, and compute the average response length on four selected benchmarks: Math-500, MinervaMath, OlympiadBench, and AIME.
As illustrated in Figure~\ref{fig:length-bs128}, compared to GRPO, our \name achieves a 17.1\% reduction in token usage on MATH, 22.3\% on MinervaMath, 18.0\% on OlympiadBench, and 15.3\% on AIME. On average, \name demonstrates a \textbf{18.1\%} decrease in token usage across the four benchmarks compared to GRPO, showcasing its superior efficiency. TreeRPO not only demonstrates an advantage in token efficiency on Qwen2.5-Math-1.5B, but also shows an efficiency advantage on Qwen2.5-Math-7B, with an average token length that is also shorter than the GRPO baseline. We show the response case of a simple question in Figure~\ref{fig:simple_case}. It can be seen that in this simple case, \name's response is more concise


\begin{figure}[htbp] 
    \centering
    \includegraphics[width=0.99\linewidth]{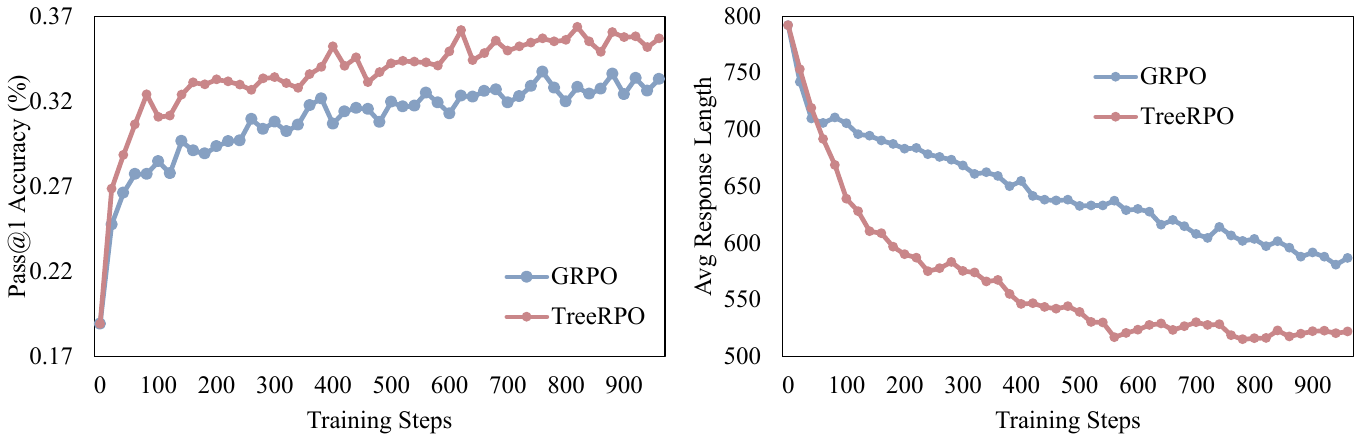}
    \caption{Comparison of \name and GRPO with $bsz = 16$. The pass@1 accuracy and the response length are calculated by taking the average on the four tested benchmarks.}\label{fig:bs16}
\end{figure}

\paragraph{The performance of \name under different hyperparameters.}
In the experiments, we conduct experimental analyses using different batch sizes, and the results are shown in Figure~\ref{fig:bs16}.
For GRPO and \name, the batch size $bsz=16/128$ has very little influence on the final performance. Our \name significantly outperforms GRPO in both two Settings. This fully demonstrates that our TreeRPO algorithm significantly outperforms the GRPO-baseline across different hyperparameters.



\section{Conclusion}
In this paper, we propose \name, which conducts tree sampling to construct step-level groups based on vanilla GRPO.
\name obtains the reward of the current step by estimating the reward of the subsequent sampling paths of the current step. This is a method that can obtain dense reward signals without the need for process reward models (PRMs). The experimental results show that \name demonstrates both effectiveness and efficiency. In the future, we will continuously improve the algorithm based on the current version and expand the scale of LLM training.

\section{Reproducibility Statement}
To ensure the reproducibility of our research, we have meticulously assembled a comprehensive reproducibility package as part of our supplementary materials. This package is designed to enable the seamless replication of all experiments detailed in our paper. It encompasses anonymized source code that implements the proposed model and training procedures. Additionally, we have included precise configuration files and scripts that specify all hyperparameters and the training commands necessary to reproduce our results.

\input{iclr2026_conference.bbl}
\bibliographystyle{iclr2026_conference}

\appendix

\section*{Appendix}

\section{The Use of LLMs}

In the preparation of this paper, large language models (LLMs), specifically DeepSeek-V3.1 and Gemini 2.5, were used solely for the purpose of polishing the writing. The LLM was employed after the core intellectual content—including the central ideas, theoretical formulations, algorithm designs, experimental setups, and result analyses—had been fully developed by the authors. The model's assistance was limited to rephrasing sentences for improved clarity, fluency, and conciseness. All prompts provided to the LLM contained only the authors' original text and instructions for grammatical or stylistic improvement.

\section{Future Work and Limitations}

\paragraph{Remove Redundant Steps.} \cite{yuan2023scaling} uses Rejection Sampling to collect correct reasoning paths for training LLMs. They find that the sampled redundant responses degrade the performance of LLMs. We consider that this phenomenon may also exist in RL. In vanilla GRPO, each response is treated equally, so responses with high similarity are repeatedly trained, which may cause performance disturbances. We believe that eliminating redundant rollouts can enhance performance while improving training efficiency through pruning.

\vspace{-3mm}
\paragraph{Precise Step Segmentation.} The step division of generated sequences in this article is implemented based on a specific token length. Give priority to exploring more precise step division methods. 
\vspace{-2mm}
\begin{itemize}
    \vspace{-1mm}
    \item One solution to be implemented is to add the \textbf{step special token} and train the language model to segment different steps by itself.
    \vspace{-1mm}
    \item Sampling at the tokens where branch paths are more likely to be generated~\citep{wang20258020rulehighentropyminority}.
\end{itemize}
\vspace{-2mm}
 We believe that more precise step cutting will provide more accurate fine-grained reward signals and further enhance the model's performance.

\vspace{-3mm}
\paragraph{Scaling on Larger Model Sizes.} Due to the limitations of time and GPU resources, our experiment can only report the 1.5b model for the time being. The experimental results of larger-sized models, such as 7b and 32b, will be updated in the future.

\vspace{-3mm}
\paragraph{Engineering Efficiency Optimization of Tree Sampling.} Tree sampling is time-consuming, and the tree sampling strategy implemented in this paper is not optimized from the perspective of the KV cache. We believe that the engineering optimization of tree sampling will significantly improve the efficiency of the training procedure.

\end{document}

%% file: math_commands.tex
\usepackage{amsmath,amsfonts,bm}

\def\eqref#1{equation~\ref{#1}}

\def\1{\bm{1}}

\DeclareMathAlphabet{\mathsfit}{\encodingdefault}{\sfdefault}{m}{sl}
\SetMathAlphabet{\mathsfit}{bold}{\encodingdefault}{\sfdefault}{bx}{n}













%% file: iclr2026_conference.bbl
\begin{thebibliography}{47}
\providecommand{\natexlab}[1]{#1}
\providecommand{\url}[1]{\texttt{#1}}
\expandafter\ifx\csname urlstyle\endcsname\relax
  \providecommand{\doi}[1]{doi: #1}\else
  \providecommand{\doi}{doi: \begingroup \urlstyle{rm}\Url}\fi

\bibitem[Bi et~al.(2023)Bi, Zhang, Jiang, Deng, Zheng, and Chen]{bi2023program}
Zhen Bi, Ningyu Zhang, Yinuo Jiang, Shumin Deng, Guozhou Zheng, and Huajun Chen.
\newblock When do program-of-thoughts work for reasoning?
\newblock \emph{arXiv preprint arXiv:2308.15452}, 2023.

\bibitem[Chen et~al.(2025)Chen, Zhang, Ma, Wang, Liang, Tu, Li, and Wong]{chen2025spcevolvingselfplaycritic}
Jiaqi Chen, Bang Zhang, Ruotian Ma, Peisong Wang, Xiaodan Liang, Zhaopeng Tu, Xiaolong Li, and Kwan-Yee~K. Wong.
\newblock Spc: Evolving self-play critic via adversarial games for llm reasoning, 2025.
\newblock URL \url{https://arxiv.org/abs/2504.19162}.

\bibitem[Cobbe et~al.(2021)Cobbe, Kosaraju, Bavarian, Chen, Jun, Kaiser, Plappert, Tworek, Hilton, Nakano, et~al.]{cobbe2021training}
Karl Cobbe, Vineet Kosaraju, Mohammad Bavarian, Mark Chen, Heewoo Jun, Lukasz Kaiser, Matthias Plappert, Jerry Tworek, Jacob Hilton, Reiichiro Nakano, et~al.
\newblock Training verifiers to solve math word problems.
\newblock \emph{arXiv preprint arXiv:2110.14168}, 2021.

\bibitem[DeepMind(2024)]{gemini-thinking}
Google DeepMind.
\newblock Gemini 2.0 flash thinking, 2024.
\newblock URL \url{https://deepmind.google/technologies/gemini/flash-thinking/}.

\bibitem[Guo et~al.(2025{\natexlab{a}})Guo, Yang, Zhang, Song, Zhang, Xu, Zhu, Ma, Wang, Bi, et~al.]{guo2025deepseek}
Daya Guo, Dejian Yang, Haowei Zhang, Junxiao Song, Ruoyu Zhang, Runxin Xu, Qihao Zhu, Shirong Ma, Peiyi Wang, Xiao Bi, et~al.
\newblock Deepseek-r1: Incentivizing reasoning capability in llms via reinforcement learning.
\newblock \emph{arXiv preprint arXiv:2501.12948}, 2025{\natexlab{a}}.

\bibitem[Guo et~al.(2025{\natexlab{b}})Guo, Xu, Liu, Ye, and Qiu]{guo2025segmentpolicyoptimizationeffective}
Yiran Guo, Lijie Xu, Jie Liu, Dan Ye, and Shuang Qiu.
\newblock Segment policy optimization: Effective segment-level credit assignment in rl for large language models, 2025{\natexlab{b}}.
\newblock URL \url{https://arxiv.org/abs/2505.23564}.

\bibitem[He et~al.(2024)He, Luo, Bai, Hu, Thai, Shen, Hu, Han, Huang, Zhang, Liu, Qi, Liu, and Sun]{olympiadbench}
Chaoqun He, Renjie Luo, Yuzhuo Bai, Shengding Hu, Zhen Thai, Junhao Shen, Jinyi Hu, Xu~Han, Yujie Huang, Yuxiang Zhang, Jie Liu, Lei Qi, Zhiyuan Liu, and Maosong Sun.
\newblock {O}lympiad{B}ench: A challenging benchmark for promoting {AGI} with olympiad-level bilingual multimodal scientific problems.
\newblock In Lun-Wei Ku, Andre Martins, and Vivek Srikumar (eds.), \emph{Proceedings of the 62nd Annual Meeting of the Association for Computational Linguistics (Volume 1: Long Papers)}, pp.\  3828--3850, Bangkok, Thailand, August 2024. Association for Computational Linguistics.
\newblock \doi{10.18653/v1/2024.acl-long.211}.
\newblock URL \url{https://aclanthology.org/2024.acl-long.211/}.

\bibitem[Hendrycks et~al.(2021)Hendrycks, Burns, Kadavath, Arora, Basart, Tang, Song, and Steinhardt]{hendrycksmath2021}
Dan Hendrycks, Collin Burns, Saurav Kadavath, Akul Arora, Steven Basart, Eric Tang, Dawn Song, and Jacob Steinhardt.
\newblock Measuring mathematical problem solving with the math dataset.
\newblock \emph{NeurIPS}, 2021.

\bibitem[Hu et~al.(2025)Hu, Zhang, Han, Jiang, Zhang, and Shum]{hu2025openreasonerzeroopensourceapproach}
Jingcheng Hu, Yinmin Zhang, Qi~Han, Daxin Jiang, Xiangyu Zhang, and Heung-Yeung Shum.
\newblock Open-reasoner-zero: An open source approach to scaling up reinforcement learning on the base model, 2025.
\newblock URL \url{https://arxiv.org/abs/2503.24290}.

\bibitem[Huang et~al.(2024)Huang, Zeng, Dai, Luo, Weng, Qing, Cui, Guo, and Zhang]{efficode}
Dong Huang, Guangtao Zeng, Jianbo Dai, Meng Luo, Han Weng, Yuhao Qing, Heming Cui, Zhijiang Guo, and Jie~M Zhang.
\newblock Effi-code: Unleashing code efficiency in language models.
\newblock \emph{arXiv preprint arXiv:2410.10209}, 2024.

\bibitem[Kazemnejad et~al.(2025)Kazemnejad, Aghajohari, Portelance, Sordoni, Reddy, Courville, and Roux]{kazemnejad2025vineppo}
Amirhossein Kazemnejad, Milad Aghajohari, Eva Portelance, Alessandro Sordoni, Siva Reddy, Aaron Courville, and Nicolas~Le Roux.
\newblock Vine{PPO}: Refining credit assignment in {RL} training of {LLM}s.
\newblock In \emph{Forty-second International Conference on Machine Learning}, 2025.
\newblock URL \url{https://openreview.net/forum?id=Myx2kJFzAn}.

\bibitem[Kwon et~al.(2023)Kwon, Li, Zhuang, Sheng, Zheng, Yu, Gonzalez, Zhang, and Stoica]{vllm}
Woosuk Kwon, Zhuohan Li, Siyuan Zhuang, Ying Sheng, Lianmin Zheng, Cody~Hao Yu, Joseph~E. Gonzalez, Hao Zhang, and Ion Stoica.
\newblock Efficient memory management for large language model serving with pagedattention.
\newblock In \emph{Proceedings of the ACM SIGOPS 29th Symposium on Operating Systems Principles}, 2023.

\bibitem[Lai et~al.(2024)Lai, Tian, Chen, Yang, Peng, and Jia]{stepdpo}
Xin Lai, Zhuotao Tian, Yukang Chen, Senqiao Yang, Xiangru Peng, and Jiaya Jia.
\newblock Step-dpo: Step-wise preference optimization for long-chain reasoning of llms, 2024.
\newblock URL \url{https://arxiv.org/abs/2406.18629}.

\bibitem[Lewkowycz et~al.(2022)Lewkowycz, Andreassen, Dohan, Dyer, Michalewski, Ramasesh, Slone, Anil, Schlag, Gutman-Solo, Wu, Neyshabur, Gur-Ari, and Misra]{minerva}
Aitor Lewkowycz, Anders Andreassen, David Dohan, Ethan Dyer, Henryk Michalewski, Vinay Ramasesh, Ambrose Slone, Cem Anil, Imanol Schlag, Theo Gutman-Solo, Yuhuai Wu, Behnam Neyshabur, Guy Gur-Ari, and Vedant Misra.
\newblock Solving quantitative reasoning problems with language models.
\newblock In S.~Koyejo, S.~Mohamed, A.~Agarwal, D.~Belgrave, K.~Cho, and A.~Oh (eds.), \emph{Advances in Neural Information Processing Systems}, volume~35, pp.\  3843--3857. Curran Associates, Inc., 2022.
\newblock URL \url{https://proceedings.neurips.cc/paper_files/paper/2022/file/18abbeef8cfe9203fdf9053c9c4fe191-Paper-Conference.pdf}.

\bibitem[Lightman et~al.(2023)Lightman, Kosaraju, Burda, Edwards, Baker, Lee, Leike, Schulman, Sutskever, and Cobbe]{verify_stepbystep}
Hunter Lightman, Vineet Kosaraju, Yura Burda, Harri Edwards, Bowen Baker, Teddy Lee, Jan Leike, John Schulman, Ilya Sutskever, and Karl Cobbe.
\newblock Let's verify step by step.
\newblock \emph{arXiv preprint arXiv:2305.20050}, 2023.

\bibitem[Lu et~al.(2024{\natexlab{a}})Lu, Dou, Wang, Cao, Dai, Feng, and Guo]{autopsv}
Jianqiao Lu, Zhiyang Dou, Hongru Wang, Zeyu Cao, Jianbo Dai, Yunlong Feng, and Zhijiang Guo.
\newblock Autopsv: Automated process-supervised verifier.
\newblock In Amir Globersons, Lester Mackey, Danielle Belgrave, Angela Fan, Ulrich Paquet, Jakub~M. Tomczak, and Cheng Zhang (eds.), \emph{Advances in Neural Information Processing Systems 38: Annual Conference on Neural Information Processing Systems 2024, NeurIPS 2024, Vancouver, BC, Canada, December 10 - 15, 2024}, 2024{\natexlab{a}}.
\newblock URL \url{http://papers.nips.cc/paper\_files/paper/2024/hash/9246aa822579d9b29a140ecdac36ad60-Abstract-Conference.html}.

\bibitem[Lu et~al.(2024{\natexlab{b}})Lu, Zhou, Ren, Wang, Shi, Pan, Zhan, and Li]{mathgenie}
Zimu Lu, Aojun Zhou, Houxing Ren, Ke~Wang, Weikang Shi, Junting Pan, Mingjie Zhan, and Hongsheng Li.
\newblock {M}ath{G}enie: Generating synthetic data with question back-translation for enhancing mathematical reasoning of {LLM}s.
\newblock In Lun-Wei Ku, Andre Martins, and Vivek Srikumar (eds.), \emph{Proceedings of the 62nd Annual Meeting of the Association for Computational Linguistics (Volume 1: Long Papers)}, pp.\  2732--2747, Bangkok, Thailand, August 2024{\natexlab{b}}. Association for Computational Linguistics.
\newblock \doi{10.18653/v1/2024.acl-long.151}.
\newblock URL \url{https://aclanthology.org/2024.acl-long.151/}.

\bibitem[Luo et~al.(2025{\natexlab{a}})Luo, Tan, Huang, Patel, Ariyak, Wu, Shi, Xin, Cai, Weber, Zhang, Li, Popa, and Stoica]{deepcoder2025}
Michael Luo, Sijun Tan, Roy Huang, Ameen Patel, Alpay Ariyak, Qingyang Wu, Xiaoxiang Shi, Rachel Xin, Colin Cai, Maurice Weber, Ce~Zhang, Li~Erran Li, Raluca~Ada Popa, and Ion Stoica.
\newblock Deepcoder: A fully open-source 14b coder at o3-mini level, 2025{\natexlab{a}}.
\newblock Notion Blog.

\bibitem[Luo et~al.(2025{\natexlab{b}})Luo, Tan, Wong, Shi, Tang, Roongta, Cai, Luo, Li, Popa, and Stoica]{deepscaler2025}
Michael Luo, Sijun Tan, Justin Wong, Xiaoxiang Shi, William~Y. Tang, Manan Roongta, Colin Cai, Jeffrey Luo, Li~Erran Li, Raluca~Ada Popa, and Ion Stoica.
\newblock Deepscaler: Surpassing o1-preview with a 1.5b model by scaling rl, 2025{\natexlab{b}}.
\newblock Notion Blog.

\bibitem[OpenAI(2024)]{o1}
OpenAI.
\newblock Learning to reason with llms, 2024.
\newblock URL \url{https://openai.com/index/learning-to-reason-with-llms/}.

\bibitem[Ouyang et~al.(2022)Ouyang, Wu, Jiang, Almeida, Wainwright, Mishkin, Zhang, Agarwal, Slama, Ray, et~al.]{rlhf}
Long Ouyang, Jeffrey Wu, Xu~Jiang, Diogo Almeida, Carroll Wainwright, Pamela Mishkin, Chong Zhang, Sandhini Agarwal, Katarina Slama, Alex Ray, et~al.
\newblock Training language models to follow instructions with human feedback.
\newblock \emph{Advances in neural information processing systems}, 35:\penalty0 27730--27744, 2022.

\bibitem[Qwen(2024)]{qwq}
Qwen.
\newblock Qwq-32b: Embracing the power of reinforcement learning, 2024.
\newblock URL \url{https://qwenlm.github.io/blog/qwq-32b/}.

\bibitem[Rafailov et~al.(2023)Rafailov, Sharma, Mitchell, Manning, Ermon, and Finn]{dpo}
Rafael Rafailov, Archit Sharma, Eric Mitchell, Christopher~D Manning, Stefano Ermon, and Chelsea Finn.
\newblock Direct preference optimization: Your language model is secretly a reward model.
\newblock \emph{Advances in Neural Information Processing Systems}, 36:\penalty0 53728--53741, 2023.

\bibitem[Schulman et~al.(2017)Schulman, Wolski, Dhariwal, Radford, and Klimov]{ppo}
John Schulman, Filip Wolski, Prafulla Dhariwal, Alec Radford, and Oleg Klimov.
\newblock Proximal policy optimization algorithms.
\newblock \emph{arXiv preprint arXiv:1707.06347}, 2017.

\bibitem[Shao et~al.(2024)Shao, Wang, Zhu, Xu, Song, Bi, Zhang, Zhang, Li, Wu, and Guo]{deepseekmathgrpo}
Zhihong Shao, Peiyi Wang, Qihao Zhu, Runxin Xu, Junxiao Song, Xiao Bi, Haowei Zhang, Mingchuan Zhang, Y.~K. Li, Y.~Wu, and Daya Guo.
\newblock Deepseekmath: Pushing the limits of mathematical reasoning in open language models, 2024.
\newblock URL \url{https://arxiv.org/abs/2402.03300}.

\bibitem[Shen et~al.(2025)Shen, Liu, Wu, Zhu, Yang, Xin, Yue, and Yan]{rlhf-sw}
Wei Shen, Guanlin Liu, Zheng Wu, Ruofei Zhu, Qingping Yang, Chao Xin, Yu~Yue, and Lin Yan.
\newblock Exploring data scaling trends and effects in reinforcement learning from human feedback.
\newblock \emph{arXiv preprint arXiv:2503.22230}, 2025.

\bibitem[Sheng et~al.(2024)Sheng, Zhang, Ye, Wu, Zhang, Zhang, Peng, Lin, and Wu]{verl}
Guangming Sheng, Chi Zhang, Zilingfeng Ye, Xibin Wu, Wang Zhang, Ru~Zhang, Yanghua Peng, Haibin Lin, and Chuan Wu.
\newblock Hybridflow: A flexible and efficient rlhf framework.
\newblock \emph{arXiv preprint arXiv: 2409.19256}, 2024.

\bibitem[Team et~al.(2025)Team, Du, Gao, Xing, Jiang, Chen, Li, Xiao, Du, Liao, et~al.]{k1.5}
Kimi Team, Angang Du, Bofei Gao, Bowei Xing, Changjiu Jiang, Cheng Chen, Cheng Li, Chenjun Xiao, Chenzhuang Du, Chonghua Liao, et~al.
\newblock Kimi k1. 5: Scaling reinforcement learning with llms.
\newblock \emph{arXiv preprint arXiv:2501.12599}, 2025.

\bibitem[Tong et~al.(2024)Tong, Zhang, Wang, Wu, and He]{tong2024dartmath}
Yuxuan Tong, Xiwen Zhang, Rui Wang, Ruidong Wu, and Junxian He.
\newblock {DART}-math: Difficulty-aware rejection tuning for mathematical problem-solving.
\newblock In \emph{The Thirty-eighth Annual Conference on Neural Information Processing Systems}, 2024.
\newblock URL \url{https://openreview.net/forum?id=zLU21oQjD5}.

\bibitem[Wang et~al.(2024)Wang, Li, Shao, Xu, Dai, Li, Chen, Wu, and Sui]{mathshepherd}
Peiyi Wang, Lei Li, Zhihong Shao, Runxin Xu, Damai Dai, Yifei Li, Deli Chen, Yu~Wu, and Zhifang Sui.
\newblock Math-shepherd: Verify and reinforce {LLM}s step-by-step without human annotations.
\newblock In Lun-Wei Ku, Andre Martins, and Vivek Srikumar (eds.), \emph{Proceedings of the 62nd Annual Meeting of the Association for Computational Linguistics (Volume 1: Long Papers)}, pp.\  9426--9439, Bangkok, Thailand, August 2024. Association for Computational Linguistics.
\newblock \doi{10.18653/v1/2024.acl-long.510}.
\newblock URL \url{https://aclanthology.org/2024.acl-long.510/}.

\bibitem[Wang et~al.(2025)Wang, Yu, Gao, Zheng, Liu, Lu, Dang, Chen, Yang, Zhang, Liu, Yang, Zhao, Yue, Song, Yu, Huang, and Lin]{wang20258020rulehighentropyminority}
Shenzhi Wang, Le~Yu, Chang Gao, Chujie Zheng, Shixuan Liu, Rui Lu, Kai Dang, Xionghui Chen, Jianxin Yang, Zhenru Zhang, Yuqiong Liu, An~Yang, Andrew Zhao, Yang Yue, Shiji Song, Bowen Yu, Gao Huang, and Junyang Lin.
\newblock Beyond the 80/20 rule: High-entropy minority tokens drive effective reinforcement learning for llm reasoning, 2025.
\newblock URL \url{https://arxiv.org/abs/2506.01939}.

\bibitem[Wei et~al.(2022)Wei, Wang, Schuurmans, Bosma, Xia, Chi, Le, Zhou, et~al.]{wei2022cot}
Jason Wei, Xuezhi Wang, Dale Schuurmans, Maarten Bosma, Fei Xia, Ed~Chi, Quoc~V Le, Denny Zhou, et~al.
\newblock Chain-of-thought prompting elicits reasoning in large language models.
\newblock \emph{Advances in neural information processing systems}, 35:\penalty0 24824--24837, 2022.

\bibitem[XAI(2024)]{grok3}
XAI.
\newblock Grok 3 beta — the age of reasoning agents, 2024.
\newblock URL \url{https://x.ai/news/grok-3}.

\bibitem[Xiang et~al.(2025)Xiang, Li, Zhang, Huang, Liu, Qu, He, Chen, Yuan, Han, Xu, Li, Sachan, and Liang]{xiang2025seephysdoesseeinghelp}
Kun Xiang, Heng Li, Terry~Jingchen Zhang, Yinya Huang, Zirong Liu, Peixin Qu, Jixi He, Jiaqi Chen, Yu-Jie Yuan, Jianhua Han, Hang Xu, Hanhui Li, Mrinmaya Sachan, and Xiaodan Liang.
\newblock Seephys: Does seeing help thinking? -- benchmarking vision-based physics reasoning, 2025.
\newblock URL \url{https://arxiv.org/abs/2505.19099}.

\bibitem[Yang et~al.(2024{\natexlab{a}})Yang, Zhang, Hui, Gao, Yu, Li, Liu, Tu, Zhou, Lin, Lu, Xue, Lin, Liu, Ren, and Zhang]{qwen25math}
An~Yang, Beichen Zhang, Binyuan Hui, Bofei Gao, Bowen Yu, Chengpeng Li, Dayiheng Liu, Jianhong Tu, Jingren Zhou, Junyang Lin, Keming Lu, Mingfeng Xue, Runji Lin, Tianyu Liu, Xingzhang Ren, and Zhenru Zhang.
\newblock Qwen2.5-math technical report: Toward mathematical expert model via self-improvement, 2024{\natexlab{a}}.
\newblock URL \url{https://arxiv.org/abs/2409.12122}.

\bibitem[Yang et~al.(2022)Yang, Qin, Chen, Lin, and Liang]{logicsolver}
Zhicheng Yang, Jinghui Qin, Jiaqi Chen, Liang Lin, and Xiaodan Liang.
\newblock {L}ogic{S}olver: Towards interpretable math word problem solving with logical prompt-enhanced learning.
\newblock In Yoav Goldberg, Zornitsa Kozareva, and Yue Zhang (eds.), \emph{Findings of the Association for Computational Linguistics: EMNLP 2022}, pp.\  1--13, Abu Dhabi, United Arab Emirates, December 2022. Association for Computational Linguistics.
\newblock \doi{10.18653/v1/2022.findings-emnlp.1}.
\newblock URL \url{https://aclanthology.org/2022.findings-emnlp.1/}.

\bibitem[Yang et~al.(2024{\natexlab{b}})Yang, Huang, Xiong, Feng, Liang, Wang, and Tang]{alignedcot}
Zhicheng Yang, Yinya Huang, Jing Xiong, Liang Feng, Xiaodan Liang, Yiwei Wang, and Jing Tang.
\newblock {A}ligned{C}o{T}: Prompting large language models via native-speaking demonstrations.
\newblock In Yaser Al-Onaizan, Mohit Bansal, and Yun-Nung Chen (eds.), \emph{Findings of the Association for Computational Linguistics: EMNLP 2024}, pp.\  2857--2896, Miami, Florida, USA, November 2024{\natexlab{b}}. Association for Computational Linguistics.
\newblock \doi{10.18653/v1/2024.findings-emnlp.163}.
\newblock URL \url{https://aclanthology.org/2024.findings-emnlp.163/}.

\bibitem[Yang et~al.(2025)Yang, Wang, Huang, Guo, Shi, Han, Feng, Song, Liang, and Tang]{yang2025optibench}
Zhicheng Yang, Yiwei Wang, Yinya Huang, Zhijiang Guo, Wei Shi, Xiongwei Han, Liang Feng, Linqi Song, Xiaodan Liang, and Jing Tang.
\newblock Optibench meets resocratic: Measure and improve {LLM}s for optimization modeling.
\newblock In \emph{The Thirteenth International Conference on Learning Representations}, 2025.
\newblock URL \url{https://openreview.net/forum?id=fsDZwS49uY}.

\bibitem[Yao et~al.(2023)Yao, Yu, Zhao, Shafran, Griffiths, Cao, and Narasimhan]{treeofthoughts}
Shunyu Yao, Dian Yu, Jeffrey Zhao, Izhak Shafran, Tom Griffiths, Yuan Cao, and Karthik Narasimhan.
\newblock Tree of thoughts: Deliberate problem solving with large language models.
\newblock In A.~Oh, T.~Naumann, A.~Globerson, K.~Saenko, M.~Hardt, and S.~Levine (eds.), \emph{Advances in Neural Information Processing Systems}, volume~36, pp.\  11809--11822. Curran Associates, Inc., 2023.
\newblock URL \url{https://proceedings.neurips.cc/paper_files/paper/2023/file/271db9922b8d1f4dd7aaef84ed5ac703-Paper-Conference.pdf}.

\bibitem[Yu et~al.(2023)Yu, Gao, and Wang]{ovm}
Fei Yu, Anningzhe Gao, and Benyou Wang.
\newblock Outcome-supervised verifiers for planning in mathematical reasoning.
\newblock \emph{arXiv preprint arXiv:2311.09724}, 2023.

\bibitem[Yu et~al.(2024)Yu, Jiang, Shi, YU, Liu, Zhang, Kwok, Li, Weller, and Liu]{yu2024metamath}
Longhui Yu, Weisen Jiang, Han Shi, Jincheng YU, Zhengying Liu, Yu~Zhang, James Kwok, Zhenguo Li, Adrian Weller, and Weiyang Liu.
\newblock Metamath: Bootstrap your own mathematical questions for large language models.
\newblock In \emph{The Twelfth International Conference on Learning Representations}, 2024.
\newblock URL \url{https://openreview.net/forum?id=N8N0hgNDRt}.

\bibitem[Yu et~al.(2025)Yu, Zhang, Zhu, Yuan, Zuo, Yue, Dai, Fan, Liu, Liu, Liu, Lin, Lin, Ma, Sheng, Tong, Zhang, Zhang, Zhang, Zhu, Zhu, Chen, Chen, Wang, Yu, Song, Wei, Zhou, Liu, Ma, Zhang, Yan, Qiao, Wu, and Wang]{dapo}
Qiying Yu, Zheng Zhang, Ruofei Zhu, Yufeng Yuan, Xiaochen Zuo, Yu~Yue, Weinan Dai, Tiantian Fan, Gaohong Liu, Lingjun Liu, Xin Liu, Haibin Lin, Zhiqi Lin, Bole Ma, Guangming Sheng, Yuxuan Tong, Chi Zhang, Mofan Zhang, Wang Zhang, Hang Zhu, Jinhua Zhu, Jiaze Chen, Jiangjie Chen, Chengyi Wang, Hongli Yu, Yuxuan Song, Xiangpeng Wei, Hao Zhou, Jingjing Liu, Wei-Ying Ma, Ya-Qin Zhang, Lin Yan, Mu~Qiao, Yonghui Wu, and Mingxuan Wang.
\newblock Dapo: An open-source llm reinforcement learning system at scale, 2025.
\newblock URL \url{https://arxiv.org/abs/2503.14476}.

\bibitem[Yuan et~al.(2023)Yuan, Yuan, Li, Dong, Lu, Tan, Zhou, and Zhou]{yuan2023scaling}
Zheng Yuan, Hongyi Yuan, Chengpeng Li, Guanting Dong, Keming Lu, Chuanqi Tan, Chang Zhou, and Jingren Zhou.
\newblock Scaling relationship on learning mathematical reasoning with large language models, 2023.

\bibitem[Yue et~al.(2025)Yue, Yuan, Yu, Zuo, Zhu, Xu, Chen, Wang, Fan, Du, Wei, Yu, Liu, Liu, Liu, Lin, Lin, Ma, Zhang, Zhang, Zhang, Zhu, Zhang, Liu, Wang, Wu, and Yan]{vapo}
Yu~Yue, Yufeng Yuan, Qiying Yu, Xiaochen Zuo, Ruofei Zhu, Wenyuan Xu, Jiaze Chen, Chengyi Wang, TianTian Fan, Zhengyin Du, Xiangpeng Wei, Xiangyu Yu, Gaohong Liu, Juncai Liu, Lingjun Liu, Haibin Lin, Zhiqi Lin, Bole Ma, Chi Zhang, Mofan Zhang, Wang Zhang, Hang Zhu, Ru~Zhang, Xin Liu, Mingxuan Wang, Yonghui Wu, and Lin Yan.
\newblock Vapo: Efficient and reliable reinforcement learning for advanced reasoning tasks, 2025.
\newblock URL \url{https://arxiv.org/abs/2504.05118}.

\bibitem[Zeng et~al.(2025)Zeng, Huang, Liu, Liu, He, Ma, and He]{simple-rl}
Weihao Zeng, Yuzhen Huang, Qian Liu, Wei Liu, Keqing He, Zejun Ma, and Junxian He.
\newblock Simplerl-zoo: Investigating and taming zero reinforcement learning for open base models in the wild, 2025.
\newblock URL \url{https://arxiv.org/abs/2503.18892}.

\bibitem[Zeng et~al.(2024)Zeng, Liu, Wan, Li, Chen, Dai, Yao, Xu, Qi, Zhao, Shen, Lu, Tan, Chen, Zhang, Shi, Wang, Guo, and Jia]{MrBen}
Zhongshen Zeng, Yinhong Liu, Yingjia Wan, Jingyao Li, Pengguang Chen, Jianbo Dai, Yuxuan Yao, Rongwu Xu, Zehan Qi, Wanru Zhao, Linling Shen, Jianqiao Lu, Haochen Tan, Yukang Chen, Hao Zhang, Zhan Shi, Bailin Wang, Zhijiang Guo, and Jiaya Jia.
\newblock Mr-ben: {A} meta-reasoning benchmark for evaluating system-2 thinking in llms.
\newblock In Amir Globersons, Lester Mackey, Danielle Belgrave, Angela Fan, Ulrich Paquet, Jakub~M. Tomczak, and Cheng Zhang (eds.), \emph{Advances in Neural Information Processing Systems 38: Annual Conference on Neural Information Processing Systems 2024, NeurIPS 2024, Vancouver, BC, Canada, December 10 - 15, 2024}, 2024.
\newblock URL \url{http://papers.nips.cc/paper\_files/paper/2024/hash/d81cb1f4dc6e13aeb45553f80b3d6837-Abstract-Conference.html}.

\bibitem[Zhang et~al.(2023)Zhang, Yang, Yuan, and Yao]{zhang2023cumulative}
Yifan Zhang, Jingqin Yang, Yang Yuan, and Andrew Chi-Chih Yao.
\newblock Cumulative reasoning with large language models.
\newblock \emph{arXiv preprint arXiv:2308.04371}, 2023.

\end{thebibliography}
